%% file: main.tex
\documentclass{article}
 \pdfoutput=1
\PassOptionsToPackage{numbers, compress}{natbib}

\usepackage[final]{neurips_distshift_2021}

\usepackage[utf8]{inputenc} 
\usepackage[T1]{fontenc}    
\usepackage[hidelinks]{hyperref}       
\usepackage{url}            
\usepackage{booktabs}       
\usepackage{amsfonts}       
\usepackage{nicefrac}       
\usepackage{microtype}      
\usepackage{xcolor}         

\usepackage{subfigure}
\usepackage{booktabs} 
\usepackage[ruled]{algorithm2e}
\usepackage{graphicx}
\usepackage{color}
\usepackage{amsmath}
\usepackage{bbm}
\usepackage{bm}
\usepackage{svg}
\usepackage{amsmath}

\newcommand{\cdashlinelr}[1]{%
  \noalign{\vskip\aboverulesep
           \global\let\@dashdrawstore\adl@draw
           \global\let\adl@draw\adl@drawiv}
  \cdashline{#1}
  \noalign{\global\let\adl@draw\@dashdrawstore
           \vskip\belowrulesep}}
           
\newcommand{\simm}{\texttt{sim}}
\newcommand{\ff}{f_{\theta}}

\title{Tackling Online One-Class Incremental Learning by Removing Negative Contrasts}

\author{%
  Nader Asadi\\
  Concordia University and Mila\\
  \texttt{nader.asadi@concordia.ca} \\
  \And
  Sudhir Mudur\\
    Concordia University\\
    \texttt{sudhir.mudur@concordia.ca} \\
  \And
  Eugene Belilovsky\\
  Concordia University and Mila\\
  \texttt{eugene.belilovsky@concordia.ca} \\
}

\begin{document}

\maketitle

\begin{abstract}
  Recent work studies the supervised online continual learning setting where a learner receives a stream of data whose class distribution changes over time.  Distinct from other continual learning settings the learner is presented new samples only once and must distinguish between all seen classes. A number of successful methods in this setting focus on storing and replaying a subset of samples alongside incoming data in a computationally efficient manner.  One recent proposal ER-AML achieved strong performance in this setting by applying an asymmetric loss based on contrastive learning to the incoming data and replayed data. However, a key ingredient of the proposed method is avoiding contrasts between incoming data and stored data, which makes it impractical for the setting where only one new class is introduced in each phase of the stream. In this work we adapt a recently proposed approach (\textit{BYOL}) from self-supervised learning to the supervised learning setting, unlocking the constraint on contrasts. We then show that supplementing this with additional regularization on class prototypes yields a new method that achieves strong performance in the one-class incremental learning setting and is competitive with the top performing methods in the multi-class incremental setting. 
\end{abstract}

\input{sections/intro}
\input{sections/method}

\input{sections/results}

\section{Conclusion}
Our major contribution is a new method to handle online one-class incremental learning without the need for negative contrasts. We demonstrated that this method can outperform strong baselines, and is also applicable and highly competitive in traditional online continual learning settings. Furthermore we have shown that recent advances in self-supervised learning without contrasts can be adapted to supervised settings, particularly in continual classification. Future work can consider if our approach can be applied in the offline one class incremental setting. 

\section*{Acknowledgements}
This research was partially funded by NSERC Discovery Grant RGPIN-2021-04104 and  RGPIN-2019-05729.
We would like to thank Concordia University and Mila for the provided computational resources and research environments.
We also would like to acknowledge the support from Compute Canada.

\bibliographystyle{plain}
\bibliography{lib.bib}

\newpage
\input{sections/appendix}

\end{document}

%% file: sections/intro.tex
\section{Introduction}
Continual learning is a paradigm that aims to allow deep learning algorithms which will have the ability to learn online from a non-stationary and never-ending stream of data. Such systems must become capable of acquiring new knowledge, while avoiding catastrophic forgetting of previously seen data, a problem commonly known and suffered by gradient-based neural networks~\cite{goodfellow2013empirical}. A number of common continual learning scenarios exist in the classification setting, each with their own set of related but also different challenges. These are often characterized along a number of axes. The first is which information is available to the learner at training and test time. In our setting we focus on the single-head setting where a learner is unaware at test time which task the data belongs to. In other words, when new classes are presented the learner must learn to distinguish them from all previously observed classes. Another common distinction is the online and offline setting. In the offline setting the learner receives the full set of data for each task and can perform unlimited training on this before moving to the next. On the other hand in the online setting the learner receives one or a small number of samples from a stream and must process and/or store these samples under a computational and memory budget. In this work we focus on the latter. 

An extreme continual classification scenario involves the learner observing individual classes at each changing point in the data stream. Until now this scenario has been studied to a limited degree in the offline setting \cite{hou2019learning, douillard2020podnet}, but has not been considered in the online setting. This is due to the inherent challenging nature of the online setting which has only recently started obtaining results competitive with iid baselines \cite{caccia2019online,caccia2021reducing}. Indeed, in \cite{douillard2020podnet} they use an expensive regularization approach that is impractical under the constraints of online continual learning. Furthermore, the method proposed by \cite{douillard2020podnet} uses a pre-trained model, performing an initial stage of offline iid pretraining on half of the classes in the dataset. Where as, our aim is to tackle the problem without any assumption of pre-trained models.

In the online continual learning setting the best performing methods rely on various forms of rehearsal, where old samples are stored in a finite memory and reused at later points in the stream. A common strong baseline is experience replay\cite{chaudhry2019tiny,aljundi2019gradient}. Recently, Caccia et al.~\cite{caccia2021reducing} showed that in online continual learning settings, after each distribution shift (task boundaries), the model observes a significant drift in the representation of previously learned classes. They hypothesize that this is fundamentally due to: (i) new class samples representations lying close to older classes and (ii) the loss structure of the standard cross entropy applied on a mix of seen and unseen classes. To mitigate this, they proposed a method to allow fine-grained control over which samples will be pushed away from other samples given an incoming batch.
The main downside of methods proposed by Caccia et al.~\cite{caccia2021reducing}, is the critical need for negative samples in the incoming batch to learn the representation of incoming data while avoiding catastrophic forgetting. The result is the inability of their methods to be applied to the more challenging setting where the model observes one class at a time in an online data stream.

In this work, we apply recent ideas from \cite{grill2020bootstrap}, which break the dependence on negative samples in self-supervised contrastive learning to the supervised continual learning setting. This allows us to maintain an asymmetric loss structure between replay samples and incoming data as in \cite{caccia2021reducing}, while providing learning on new incoming class data without the need to contrast to old classes or their representations. Augmenting this approach with a regularization term that constrains class prototypes, further yields a new method which far exceeds performance of strong online CL benchmarks in the one class incremental setting and is competitive with the top performing methods in the multi-class incremental setting.


%% file: sections/method.tex
\section{Methods}


Similar to~\cite{caccia2021reducing}, given a model $f_{\theta}(x)$ representing a  neural network architecture with parameters $\theta$, we want to minimize the classification loss $\mathcal{L}$  on the newly arriving data batch while not negatively impacting previous learning of other classes, and have the ability to be applied to one-class incremental settings. We opt for a specific loss structure on the incoming batch that would enable the model to learn the representation of each class independently and in isolation from all other classes, either in the incoming batch or in buffered samples. This removes the need for negative samples and enables the model to learn useful representations, even in one-class incremental settings. We first present our supervised modification of BYOL and apply it in the asymmetric setting of \cite{caccia2021reducing}, then propose our new method.

\subsection{Supervised BYOL (SupBYOL)}
In order to remove the need for negative samples to learn the representation of incoming batch, we apply a supervised modification of BYOL~\cite{grill2020bootstrap} on the incoming data. BYOL uses a twin architecture with online and target networks. The online network is comprised of three stages: an encoder $f_{\theta}$, a projector $g_{\theta}$, and a predictor $q_{\theta}$. The target network has the same architecture but with different parameters $\xi$, and is the exponential moving average of the online network: $\xi \leftarrow \tau\xi + (1 - \tau)\theta$.
We use the following loss, a modification of BYOL loss~\cite{grill2020bootstrap}, on the incoming data  $\mathbf{X}^{in}$.
\begin{equation}
    \mathcal{L}_{1}^{byol} = - \frac{1}{N}\sum_{\mathbf{x}_i \in \mathbf{X}_{in}}\frac{1}{|P(\mathbf{x}_i)|}\sum_{\mathbf{x}_p \in P(\mathbf{x}_i)}\simm\big(q_{\theta}(z_{\theta}), z_{\xi}\big)
\end{equation}
where $\simm(a,b) = \frac{a^Tb}{\tau\|a\|\|b\|} $, and $z_{\theta} = g_{\theta}(f_{\theta}(x_{i}))$ and $z_{\xi} = g_{\xi}(f_{\xi}(x_{i}))$ are the projections outputted by online and target networks respectively. We denote the incoming $N$ data points by $\mathbf{X}^{in}$, data replayed from the buffer by $\mathbf{X}^{bf}$, and the set of positive samples with respect to $\mathbf{x}_i$ by $P$. Following~\cite{caccia2021reducing}, for the rehearsal step, we apply a modified cross-entropy objective as per~\cite{qi2018low}.
\begin{equation}
    \mathcal{L}_2(\mathbf{X}^{bf}) = -\sum_{x_{i} \in \mathbf{X}_{bf}}\log\frac{\exp\big({\simm\big(\mathbf{c}_{y(x_{i})}, \ff\mathbf{z_{i}}\big)}\big)}{\sum_{y\in \mathcal{Y}_{all}} \exp\big(\simm\big(\mathbf{c}_y, \mathbf{z_{i}}\big)\big)}
\end{equation}

where $Y_{all}$ is the set of all observed classes, and $z_{i} = g_{\theta}(f_{\theta}(x_{i}))$.

\begin{figure*}[t!]
    \begin{minipage}{0.5\textwidth}
    \includegraphics[width=1.0\textwidth, bb=0 0 585 350]{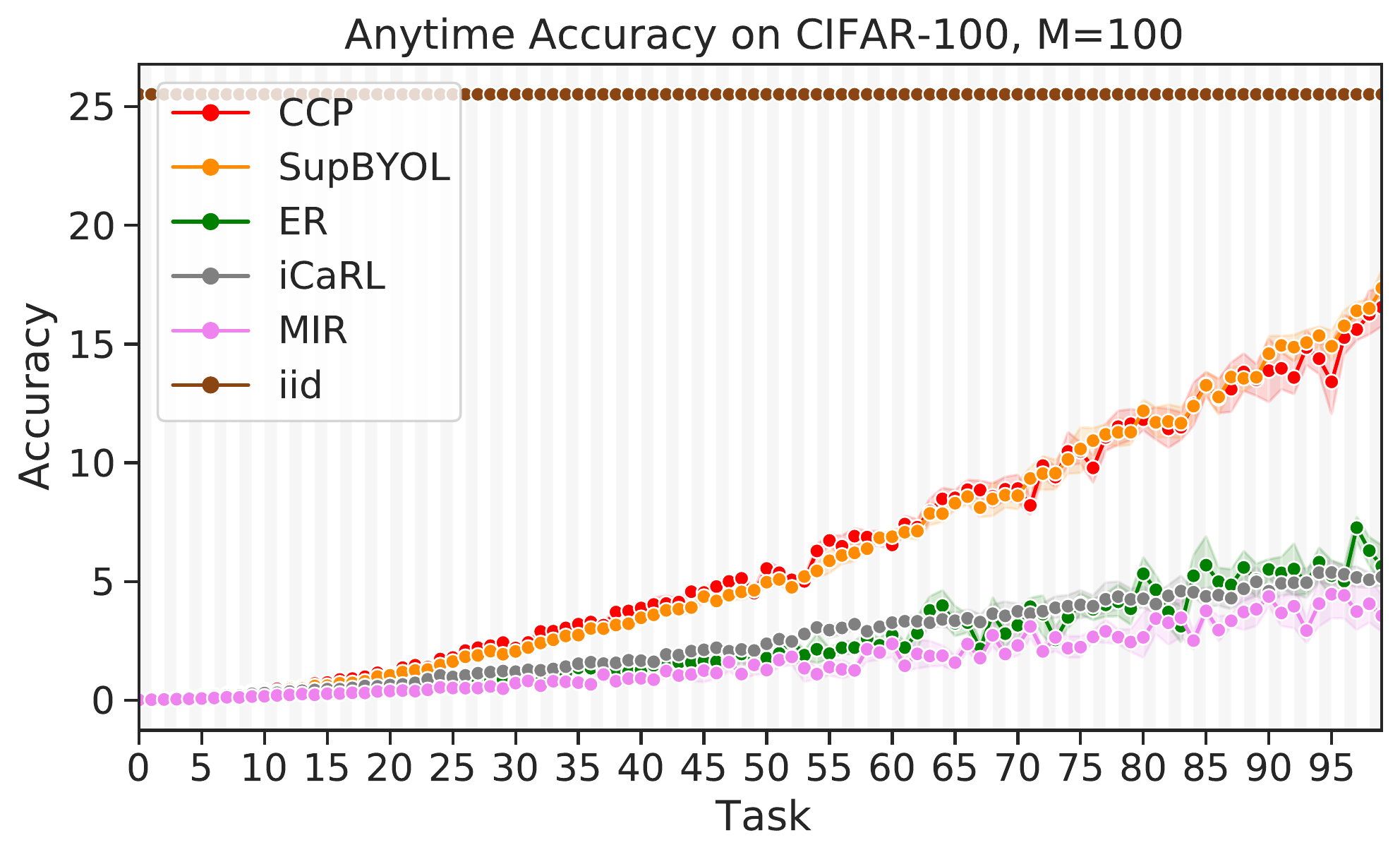}
    \end{minipage}\hfill
    \begin{minipage}{0.5\textwidth}
    \includegraphics[width=1.0\textwidth, bb=0 0 585 350]{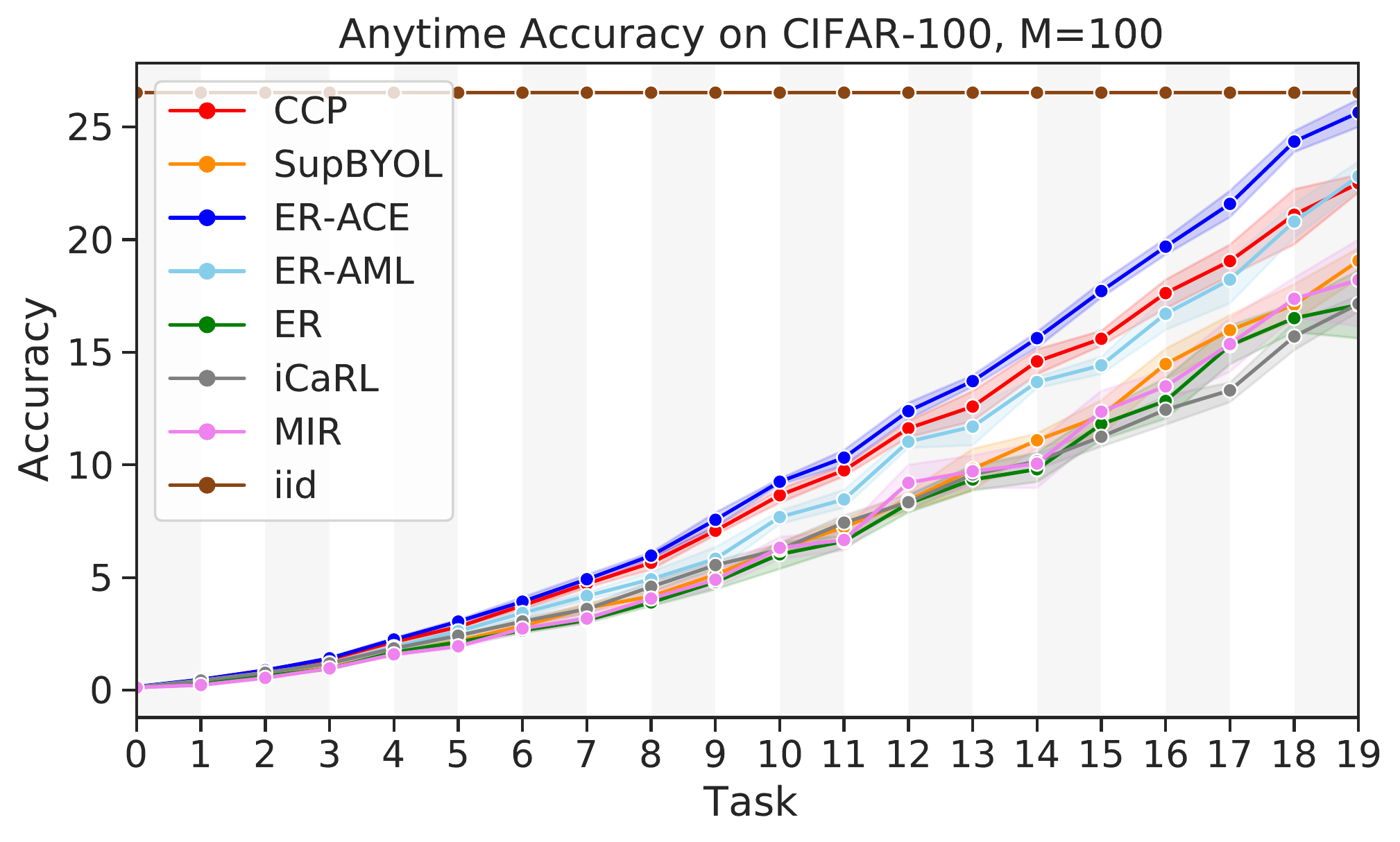}
    \end{minipage}\hfill
   \caption{\small Split CIFAR-100 anytime evaluation results in one-class(left) and multi-class(right) incremental settings with $M=100$. SupBOYL and CCP greatly outperform the existing methods in one-class setting while being competitive with the top performing methods in multi-class setting. Note that ER-ACE and ER-AML cannot be applied in one-class setting since they require more than one class in the incoming data.
   }
    \label{fig:Cifar_M20_M}
\end{figure*}

\subsection{CCP: Continual Contrast of Class Prototypes}
Although BYOL can learn the representation of each class independently from other classes, in case of supervised learning, specially where we have multiple classes in the incoming data, it is tricky to evade collapsed representations. In some cases, we observed representation collapse between new classes in the incoming data or old classes in the memory buffer, which results in forgetting of the corresponding classes (drop in accuracy). Also, due to the twin architecture design, BYOL has a larger compute footprint than ER-AML, which makes it less suitable for online continual learning.

Inspired by~\cite{caron2020unsupervised}, we use randomly initialized prototypes as class cluster representatives to reduce intra-class variance while enforcing inter-class variance.
We represent each observed class $y \in \mathcal{Y}$ by a prototype $\mathbf{c}_{y}$ in prototypes memory $\mathcal{C}$. The network parameters $\theta$ and class prototypes $\mathcal{C}$ are jointly optimized to project an instance $\mathbf{x}_i \in \mathbf{X}^{in}$ of class $y$ close to its corresponding cluster prototype $\mathbf{c}_{y}$ as well as other samples of class $y$ in the batch, i.e. positive samples $P(\mathbf{x}_i)$. In order to evade collapsed representations, we use a contrast term between class prototypes which enforces inter-class variance. So, as the prototypes act like a representative for the whole corresponding class samples, we can enforce inter-class variance without any need for negative samples, and also provide stable training without BYOL's optimization tricks such as twin architecture and predictor head. We formulate the objective as follows:

\begin{equation}\label{eqn:ccp}
    \mathcal{L}_{1}^{ccp} = -\frac{1}{N}\sum_{\mathbf{x}_i \in \mathbf{X}_{in}}\left(\simm\large(\mathbf{z}_{i}, \mathbf{c}_{y(x_{i})}\large) + \frac{1}{|P(\mathbf{x}_i)|} \sum_{\mathbf{x}_p\in \mathbf{P\mathbf(x_i)}} \simm\big(\mathbf{z}_{i}, \mathbf{z}_{p}\big)\right) + \frac{1}{|\mathcal{C}|} \sum_{\mathbf{c}_i \in \mathcal{C}} \sum_{\mathbf{c}_j \in \mathcal{C}/i}
    \simm\big(\mathbf{c}_{i}, \mathbf{c}_{j}\big)
\end{equation}

where $\simm(a,b) = \frac{a^Tb}{\tau\|a\|\|b\|} $, $z_{i} = g_{\theta}(f_{\theta}(x_{i}))$, and $y(x)$ denotes the class label of $x$. As in the case of SupBYOL, we combine this with the same $\mathcal{L}_2$ term applied to the buffered samples.
In order to avoid contrast between new and old classes introduced by the last term, we do not directly update old class prototypes using gradients. In other words, for the incoming data, we perform stochastic optimization to minimize $\mathcal{L}_{1}+\mathcal{L}_{2}$ with respect to $\theta$ and $c \in \mathcal{C}^{in}$, where $\mathcal{C}^{in}$ is the class labels of the incoming data.
However, to update the old class prototypes, we follow \cite{de2019continual} and use the replayed samples to obtain a momentum update after each training step to stabilize the model against drastic changes in the representation of learned classes: $c_{y} \leftarrow \alpha c_{y} + (1 - \alpha)\bar{c}_{y}$ where $y \in \mathcal{Y}_{bf}$ denotes the label of old classes stored in the buffer and $\bar{c}_{y}$ is the updated prototype for class $y$.

%% file: sections/results.tex
\section{Experiments}

\setlength{\tabcolsep}{2.5pt}
\begin{table*}
\small
    \centering
    \begin{tabular}{ll}
        \begin{tabular}{c c c c c}
            \hline
            &\multicolumn{3}{c}{Accuracy ($\uparrow$ is better)} \\
                     & $M=5$ & $M=20$ & $M=50$ & $M=100$ \\\hline \hline
                iid online    & 63.4 \tiny{$\pm 0.6$} & 63.4 \tiny{$\pm 0.6$} & 63.4 \tiny{$\pm 0.6$} & 63.4 \tiny{$\pm 0.6$}\\
                 \hline
                
                fine-tuning  & 10.0 \tiny{$\pm 0.0$} & 10.0 \tiny{$\pm 0.0$} & 10.0 \tiny{$\pm 0.0$} & 10.0 \tiny{$\pm 0.0$}\\
               
                ER \cite{chaudhry2019tiny} &15.7 \tiny{$\pm 1.7$}& 18.2 \tiny{$\pm 2.6$} & 19.6 \tiny{$\pm 3.1$} & 21.2 \tiny{$\pm 2.3$}\\
                
                iCarl \cite{rebuffi2017icarl}   & 18.8 \tiny{$\pm  1.6$} & 20.5 \tiny{$\pm 1.5$} & 23.3 \tiny{$\pm 1.2$} &24.4 \tiny{$\pm 0.9$}\\
                
                MIR \cite{aljundi2019online} &17.8 \tiny{$\pm 1.8$}&  19.3 \tiny{$\pm  2.7$} & 20.2 \tiny{$\pm 2.2$} & 22.0 \tiny{$\pm 2.4$} \\
                
                DER++ \cite{buzzega2020dark}&20.1 \tiny{$\pm 1.4$}& 21.2 \tiny{$\pm 2.4$} & 23.2 \tiny{$\pm 2.4$} & 24.6 \tiny{$\pm 1.5$} \\
                SupBYOL (ours) &\textbf{24.4} \tiny{$\pm 1.2$}& 30.6 \tiny{$\pm 1.4$} & 33.3 \tiny{$\pm 2.1$} & 36.0 \tiny{$\pm 2.3$} \\
                CCP (ours) &\textbf{24.2} \tiny{$\pm 0.9$}& \textbf{34.3} \tiny{$\pm 1.1$} & \textbf{36.0} \tiny{$\pm 1.3$} & \textbf{39.1}\tiny{$\pm 1.1$} \\\hline
                
            \end{tabular}
        &
            \begin{tabular}{c c c}
            \hline
            \multicolumn{3}{c}{Forgetting ($\downarrow$ is better)} \\
                 $M=20$ & $M=50$ & $M=100$  \\\hline \hline
                N/A & N/A &N/A \\\hline
                100.0\tiny{$\pm 0.0$} & 100.0\tiny{$\pm 0.0$} & 100.0\tiny{$\pm 0.0$}  \\
                54.7\tiny{$\pm 2.5$} & 51.1\tiny{$\pm 2.8$} & 45.7\tiny{$\pm 2.6$}  \\
                46.1\tiny{$\pm 1.7$} & 42.8\tiny{$\pm 1.7$} & 40.1\tiny{$\pm 1.4$}  \\
                52.0\tiny{$\pm 2.9$} & 47.4\tiny{$\pm 2.6$} & 41.9\tiny{$\pm 2.8$}  \\
                58.9\tiny{$\pm 2.2$} & 53.3\tiny{$\pm 2.2$} & 49.8\tiny{$\pm 1.8$}  \\
                28.3\tiny{$\pm 1.6$} & 25.6\tiny{$\pm 2.1$} & 23.1\tiny{$\pm 2.4$}  \\
                \textbf{27.2}\tiny{$\pm 1.0$} & \textbf{24.2}\tiny{$\pm 1.1$} & \textbf{21.9}\tiny{$\pm 1.0$}  \\ \hline
            \end{tabular}
        \end{tabular}
\caption{\small Accuracy and Forgetting results on Split CIFAR-10 in \textbf{one-class} incremental setting(10 tasks) with augmentations and different buffer sizes. Averages and standard deviations are computed over five runs. SupBYOL and CCP outperform other methods with a considerable margin in both Accuracy and Forgetting.
}
\label{tab:cifar10_er_main}
\end{table*}

\setlength{\tabcolsep}{2.5pt}
\begin{table*}
\small
    \centering
    \begin{tabular}{ll}
        \begin{tabular}{c c c c c}
            \hline
            &\multicolumn{3}{c}{Accuracy ($\uparrow$ is better)} \\
                     & $M=5$ & $M=20$ & $M=50$ & $M=100$ \\\hline \hline
                iid online    & 63.4 \tiny{$\pm 0.6$} & 63.4 \tiny{$\pm 0.6$} & 63.4 \tiny{$\pm 0.6$} & 63.4 \tiny{$\pm 0.6$}\\
                 \hline
                
                fine-tuning  & 17.9 \tiny{$\pm 0.2$}& 17.9 \tiny{$\pm 0.2$}& 17.9 \tiny{$\pm 0.2$}& 17.9 \tiny{$\pm 0.2$}\\
               
               iCarl \cite{rebuffi2017icarl}   & 33.4\tiny{$\pm 1.0$}& 39.2\tiny{$\pm 0.8$}& 41.6\tiny{$\pm 0.9$}& 42.3\tiny{$\pm 0.8$}\\
              
                ER \cite{chaudhry2019tiny} & 28.4\tiny{$\pm 1.0$}& 40.3\tiny{$\pm 0.6$}& 42.8\tiny{$\pm 1.2$}& 49.4\tiny{$\pm 1.3$}\\
                
                MIR \cite{aljundi2019online} & 29.8\tiny{$\pm 1.0$}& 41.8\tiny{$\pm 0.6$}& 45.6\tiny{$\pm 0.7$}& 49.3\tiny{$\pm 0.6$}\\
                
                DER++ \cite{buzzega2020dark}&31.8 \tiny{$\pm 0.9$}& 39.3\tiny{$\pm 1.0$}& 46.7\tiny{$\pm 1.1$}& 52.3\tiny{$\pm 1.1$}\\
                
                ER-AML \cite{caccia2021reducing}& \textbf{36.4}\tiny{$\pm 1.4$}& \textbf{47.7}\tiny{$\pm 0.7$}&\textbf{52.6} \tiny{$\pm 1.1$}& \textbf{55.7}\tiny{$\pm 1.3$}\\
                
                ER-ACE \cite{caccia2021reducing}& 35.1\tiny{$\pm 0.9$}& 43.4\tiny{$\pm 1.6$}&49.3 \tiny{$\pm 1.2$}& 53.7\tiny{$\pm 1.1$}\\
                
                SupBYOL (ours)& 25.4 \tiny{$\pm 1.2$}& 36.6 \tiny{$\pm 1.3$}& 41.4 \tiny{$\pm 1.2$}& 43.6 \tiny{$\pm 1.8$}\\
                
                CCP (ours) & 34.2 \tiny{$\pm 0.9$}& 42.0 \tiny{$\pm 1.1$}& 47.6 \tiny{$\pm 1.0$}& 51.2 \tiny{$\pm 0.9$} \\\hline
                
            \end{tabular}
        &
            \begin{tabular}{c c c}
            \hline
            \multicolumn{3}{c}{Forgetting ($\downarrow$ is better)} \\
                 $M=20$ & $M=50$ & $M=100$  \\\hline \hline
                N/A & N/A &N/A \\\hline
                80.9\tiny{$\pm 0.1$} & 80.9\tiny{$\pm 0.1$} & 80.9\tiny{$\pm 0.1$}  \\
                31.3\tiny{$\pm 0.8$} & 30.8\tiny{$\pm 1.2$} & 29.4\tiny{$\pm 1.6$}  \\
                26.8\tiny{$\pm 0.8$} & 24.4\tiny{$\pm 1.2$} & 22.8\tiny{$\pm 1.6$}  \\
                38.2\tiny{$\pm 1.2$} & 21.6\tiny{$\pm 0.9$} & 15.8\tiny{$\pm 1.1$}  \\
                29.7\tiny{$\pm 1.1$} & 24.5\tiny{$\pm 1.0$} & 19.0\tiny{$\pm 1.1$}  \\
                19.8\tiny{$\pm 0.4$} & 16.4\tiny{$\pm 0.4$} & 15.9\tiny{$\pm 0.3$}  \\
                \textbf{18.3}\tiny{$\pm 0.6$} & \textbf{15.2}\tiny{$\pm 0.8$} & \textbf{14.6}\tiny{$\pm 0.8$}  \\
                20.7\tiny{$\pm 1.1$} & 18.9\tiny{$\pm 1.2$} & 17.4\tiny{$\pm 1.2$}  \\
                19.7\tiny{$\pm 0.8$} & 16.3\tiny{$\pm 1.0$} & \textbf{14.4}\tiny{$\pm 0.8$}  \\ \hline
            \end{tabular}
        \end{tabular}
\caption{\small Accuracy and Forgetting results on Split CIFAR-10 in \textbf{multi-class} incremental setting (5 tasks) with augmentations and different buffer sizes. Averages and standard deviations are computed over five runs. Our proposed method, CCP, is competitive with top performing methods in this setting.} 
\label{tab:cifar10_er_main2}
\end{table*}

Our method enables the model to learn the representation of each class in isolation from other classes without the need to have more than one class in the incoming data. Following \cite{chaudhry2019continual, caccia2021reducing}, we use a reduced Resnet-18 with \textit{batch size} and the \textit{rehearsal batch size} of \textit{10} samples. All of the models are trained in single head setting, so the task id is not revealed to the model at test time. 
Similar to \cite{caccia2021reducing}, we find data augmentation to be useful in most of the settings, specially in simple datasets like CIFAR-10 with a small buffer size where the model might overfit on the buffer samples. 
Our data augmentation pipline consists of a simple random crop followed by random horizontal flip.
We use SGD for optimization with a learning rate of 0.1 as in \cite{aljundi2019online}. We now present the experiments on 10 and 100 task settings for Split CIFAR-10 and Split CIFAR100.

\textbf{Split CIFAR-10} typically partitions the dataset into 5 disjoint tasks containing two classes each ( \cite{aljundi2019online,shim2020online}). In this work we also consider partitioning into 10 disjoint sets (1 class each). When applying  the 10 class split we will indicate $S=1$, while the case of 2 class tasks will be denoted $S=2$.

We show in Table~\ref{tab:cifar10_er_main} results for split CIFAR10 on $S=1$ and for $S=2$ in Table~\ref{tab:cifar10_er_main2} showing the overall accuracy and forgetting at the end of the task sequence for a variety of memory settings. Observe that in the one class setting ER-ACE and ER-AML cannot be applied as they require other classes in the incoming data. SupBOYL and CCP greatly outperform the existing methods in this setting, with CCP obtaining top performance in all categories. For the multi-class setting of $S=2$, SupBYOL performs poorly, but CCP greatly improves upon SupBYOL and achieves performance close to ER-AML and ER-ACE in both accuracy and forgetting categories. 

\textbf{Split CIFAR-100} Consists of 100 classes typically split into 20 tasks, each containing a disjoint set of 5 labels ($S=5$).  In our one class incremental work we will also consider the case of splitting into 100 distinct task switches ($S=1$).   All CIFAR experiments process $32 \times32$ images. 

In Figure~\ref{fig:Cifar_M20_M}, we show the results on $S=1$ and $S=5$. In the $S=1$ setting especially as the number of classes grows, we observe increasing margins over existing methods for both SupBYOL and CCP. On the other hand, in the $S=5$ setting, SupBYOL does not perform well as in the CIFAR-10, while CCP matches the performance of ER-AML and is close to the performance of ER-ACE (which nearly matches the i.i.d performance).

%% file: sections/appendix.tex
\appendix

\section{Appendix}

In this section, we provide some extra anytime evaluation results for various memory sizes on the considered datasets , i.e. Split CIFAR-10 and Split CIFAR-100, in both one-class(left plots) and multi-class(right plots) incremental learning. The results complement the ones presented in the main text.

\begin{figure*}[h]
    \begin{minipage}{0.5\textwidth}
    \includegraphics[width=1.0\textwidth, bb=0 0 585 350]{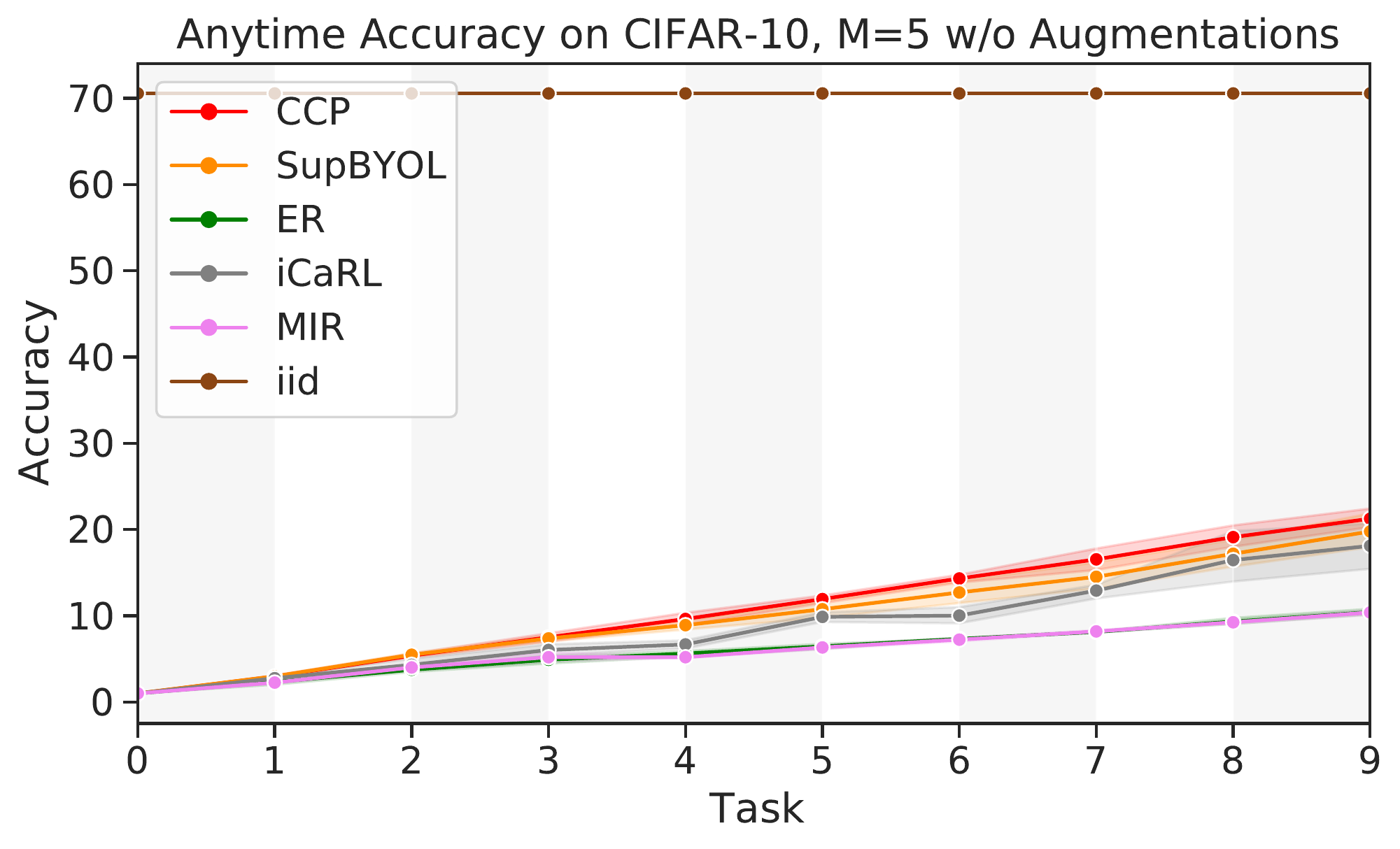}
    \end{minipage}\hfill
    \begin{minipage}{0.5\textwidth}
    \includegraphics[width=1.0\textwidth, bb=0 0 585 350]{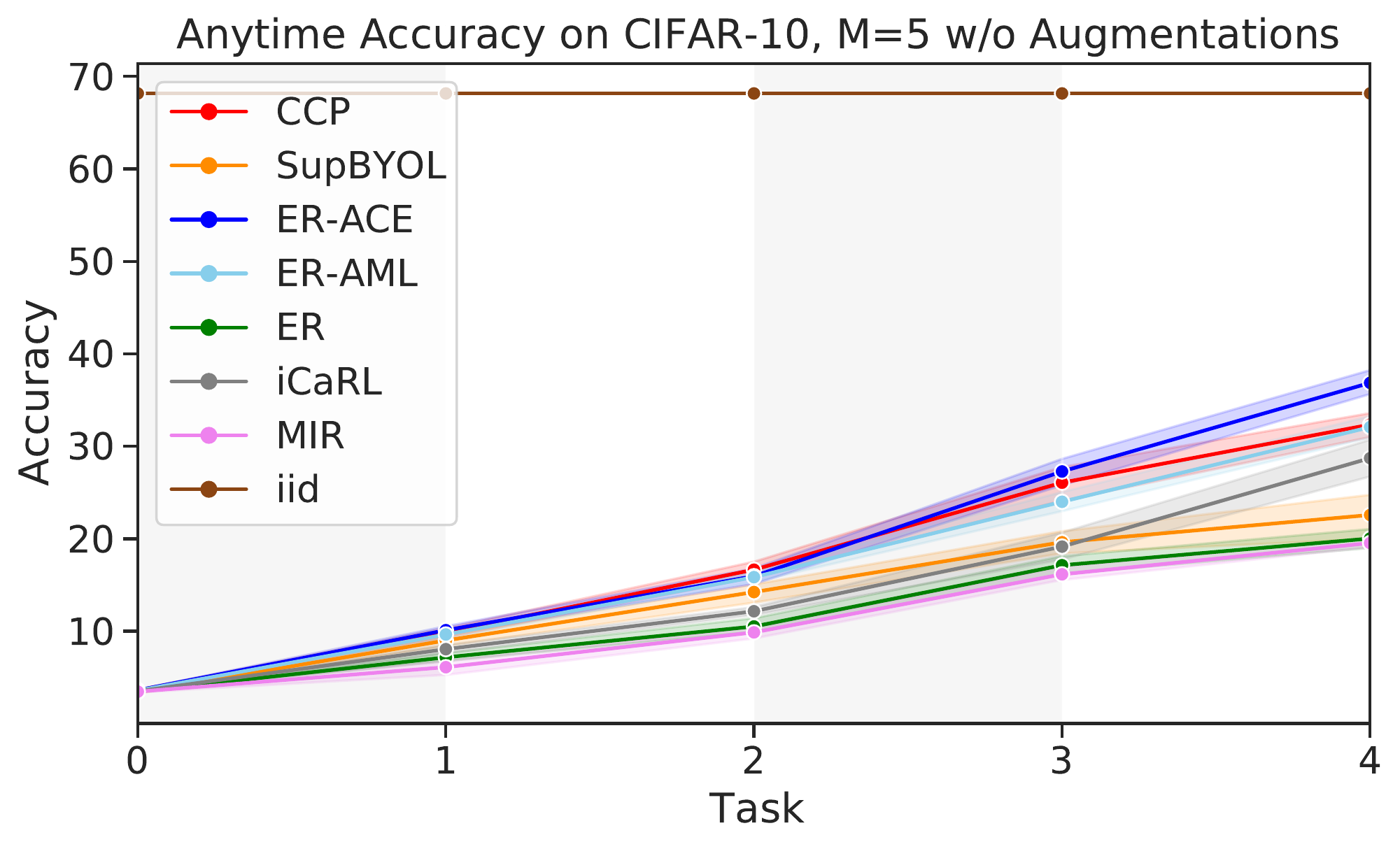}
    \end{minipage}\hfill
\end{figure*}

\vspace{10pt}

\begin{figure*}[h]
    \begin{minipage}{0.5\textwidth}
    \includegraphics[width=1.0\textwidth, bb=0 0 585 350]{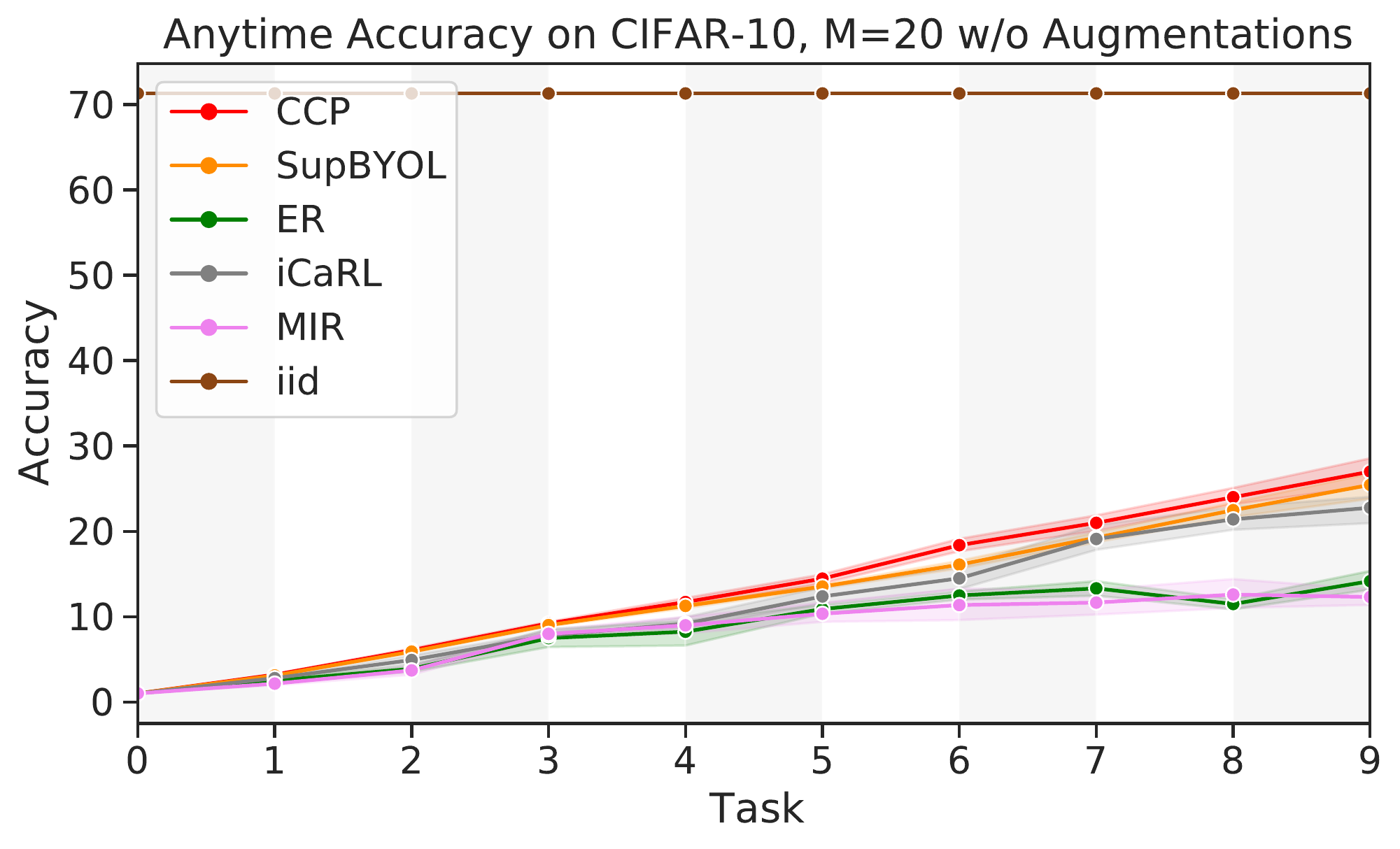}
    \end{minipage}\hfill
    \begin{minipage}{0.5\textwidth}
    \includegraphics[width=1.0\textwidth, bb=0 0 585 350]{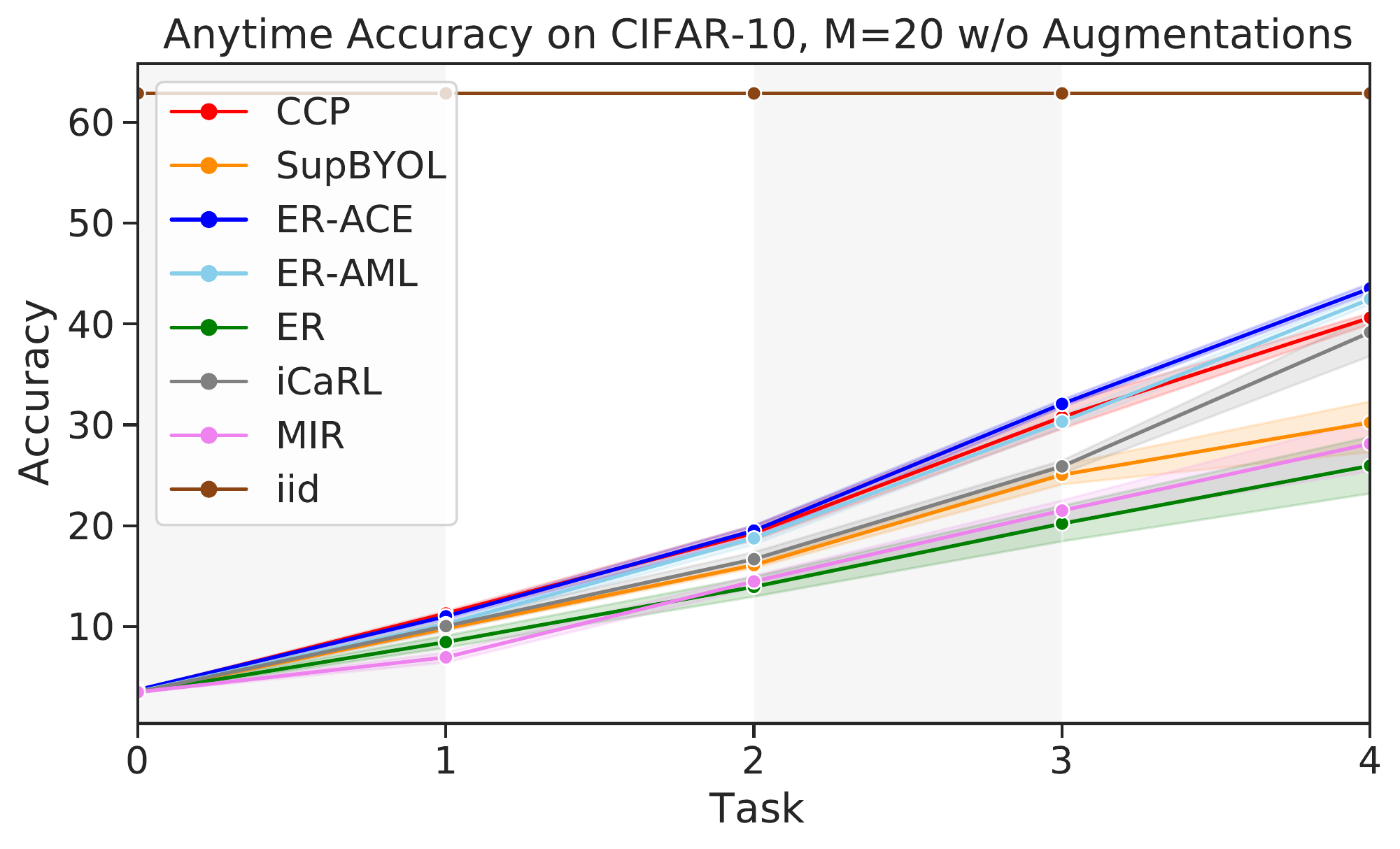}
    \end{minipage}\hfill
\end{figure*}

\vspace{10pt}

\begin{figure*}[h]
    \begin{minipage}{0.5\textwidth}
    \includegraphics[width=1.0\textwidth, bb=0 0 585 350]{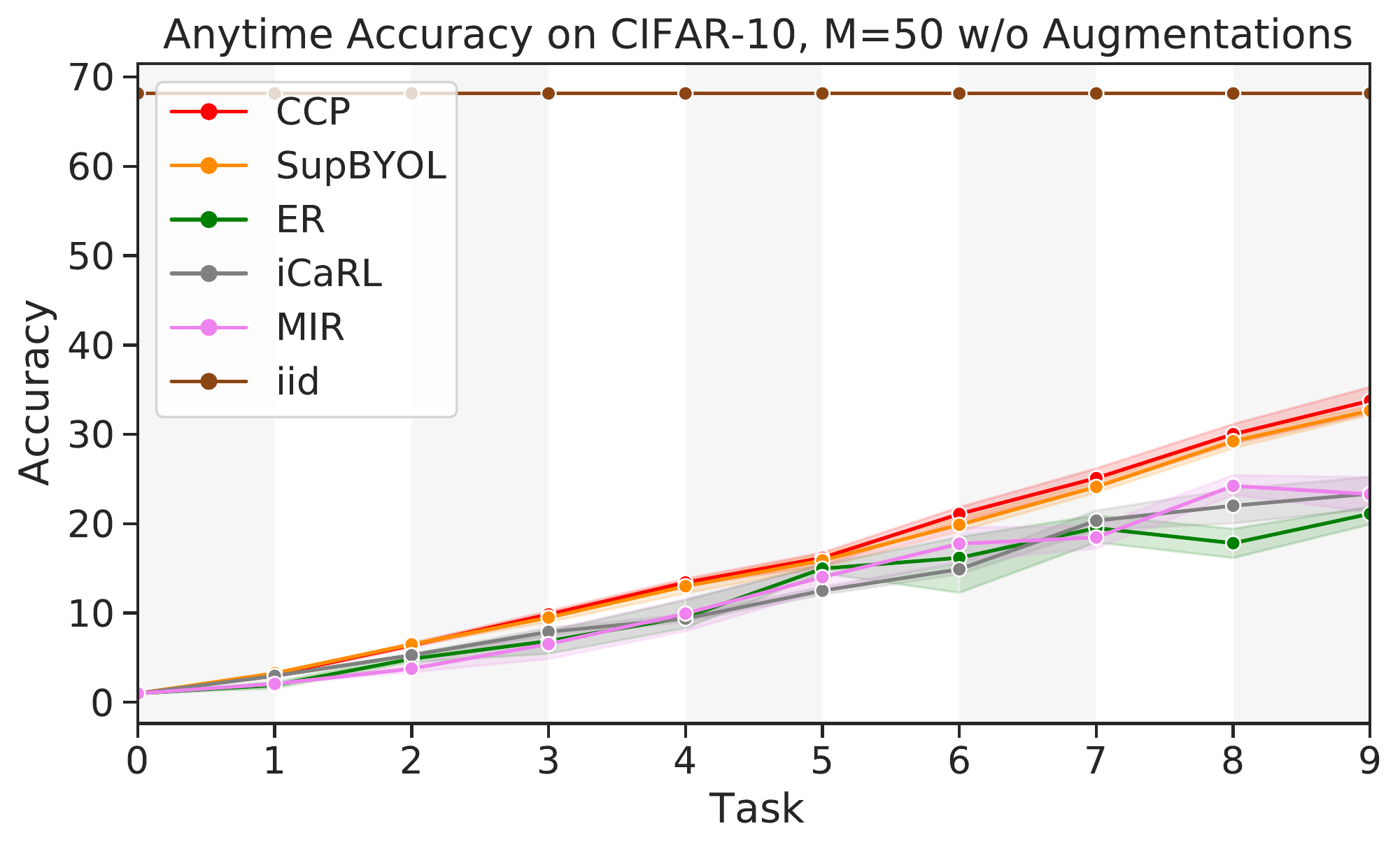}
    \end{minipage}\hfill
    \begin{minipage}{0.5\textwidth}
    \includegraphics[width=1.0\textwidth, bb=0 0 585 350]{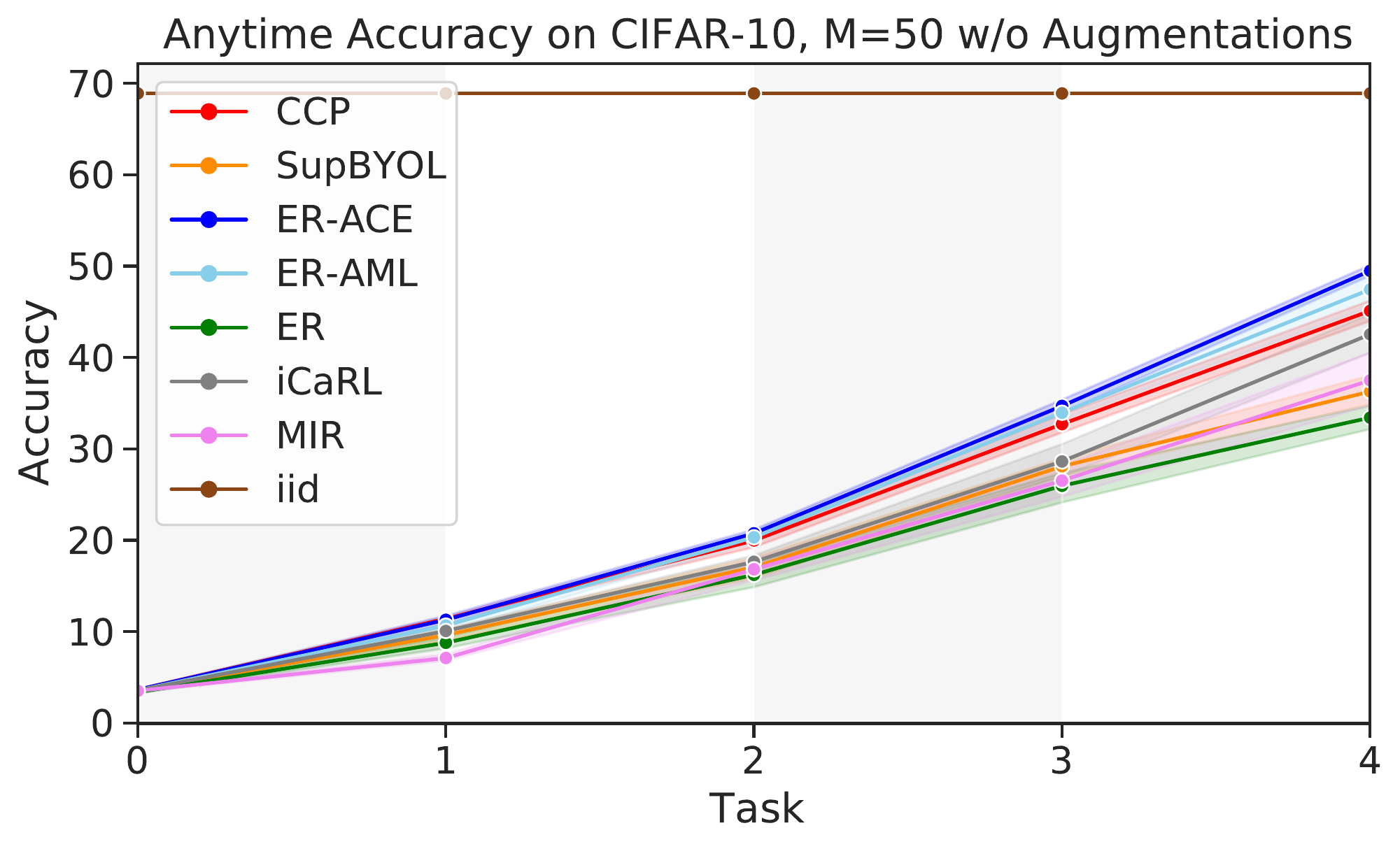}
    \end{minipage}\hfill
\end{figure*}

\begin{figure*}[h]
    \begin{minipage}{0.5\textwidth}
    \includegraphics[width=1.0\textwidth, bb=0 0 585 350]{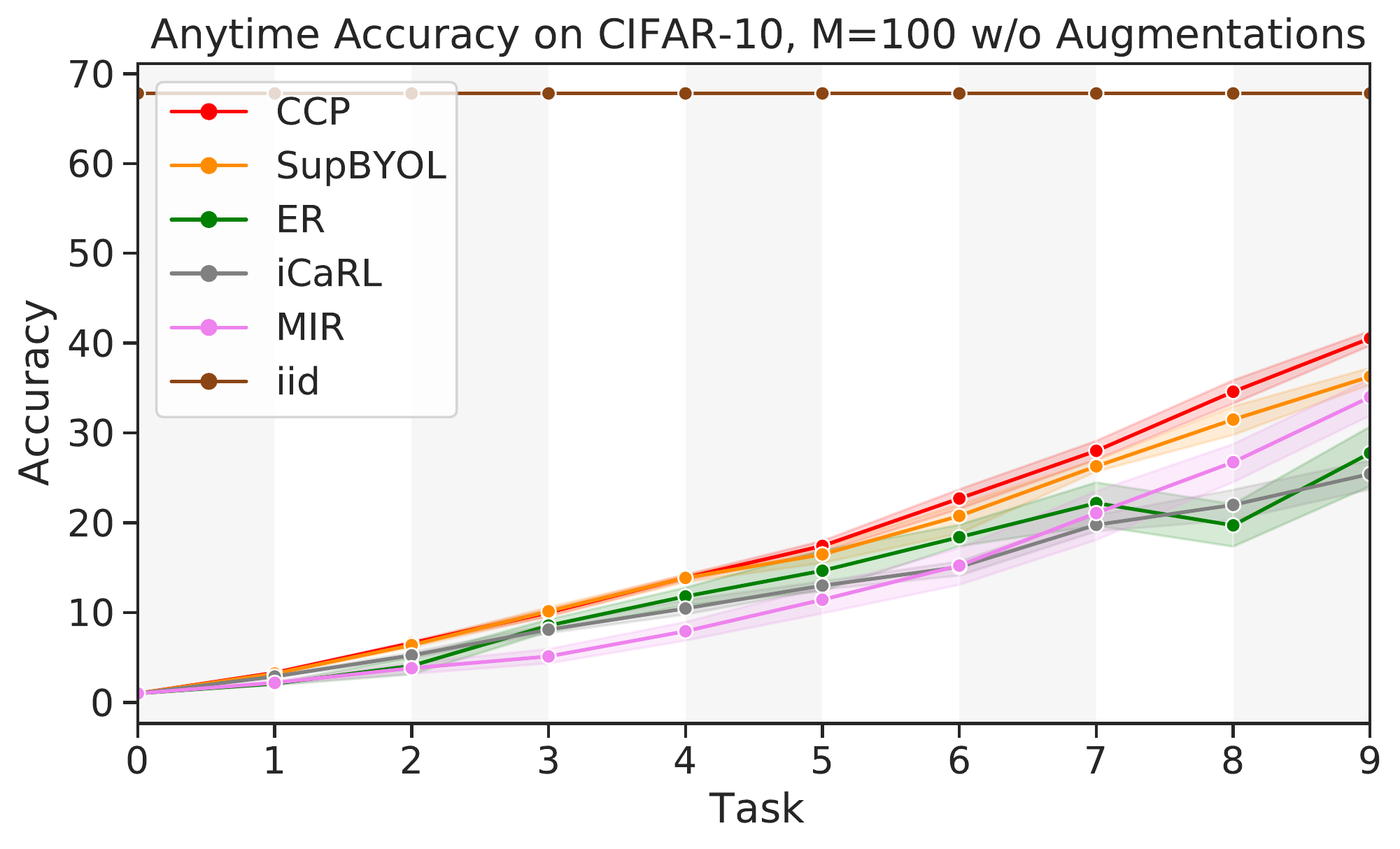}
    \end{minipage}\hfill
    \begin{minipage}{0.5\textwidth}
    \includegraphics[width=1.0\textwidth, bb=0 0 585 350]{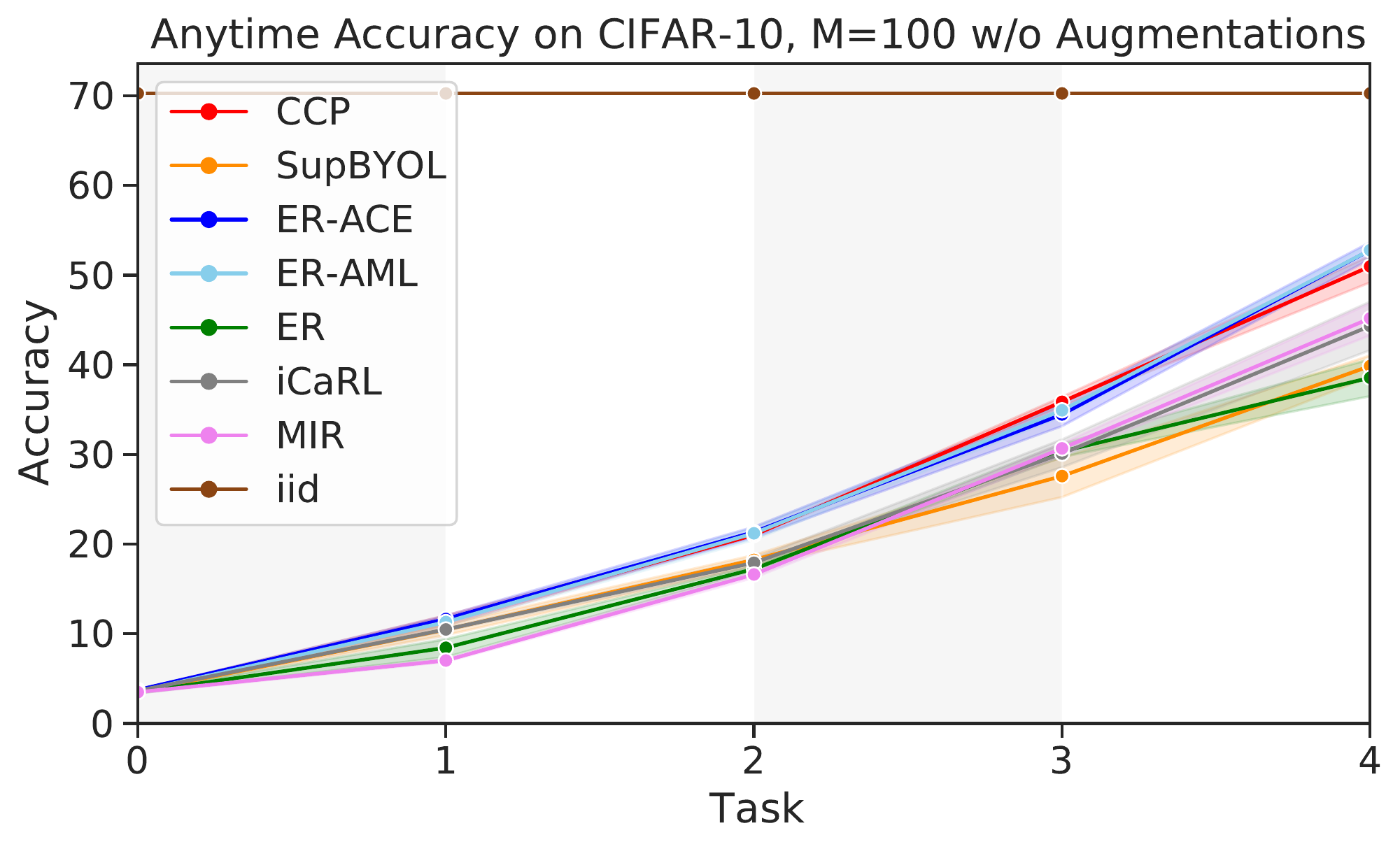}
    \end{minipage}\hfill
\end{figure*}

\begin{figure*}[h]
    \begin{minipage}{0.5\textwidth}
    \includegraphics[width=1.0\textwidth, bb=0 0 1280 640]{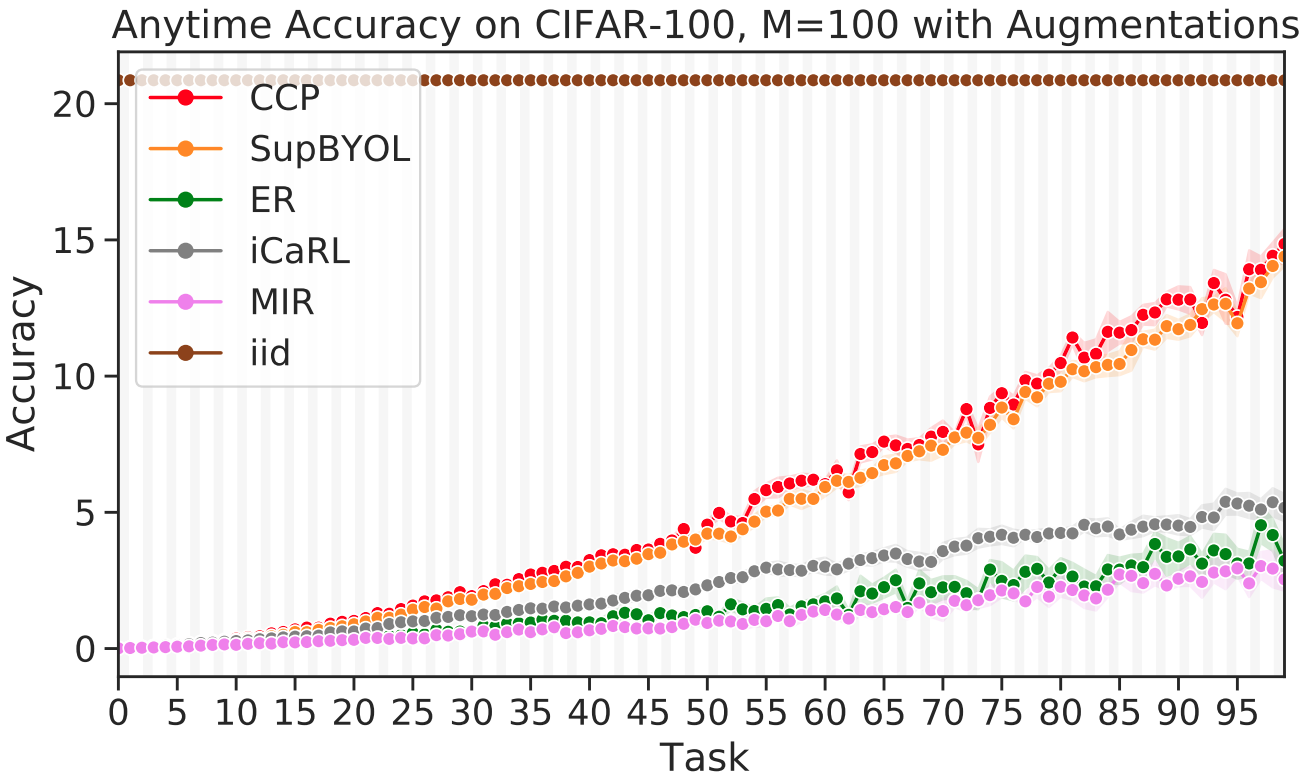}
    \end{minipage}\hfill
    \begin{minipage}{0.5\textwidth}
    \includegraphics[width=1.0\textwidth, bb=0 0 1280 640]{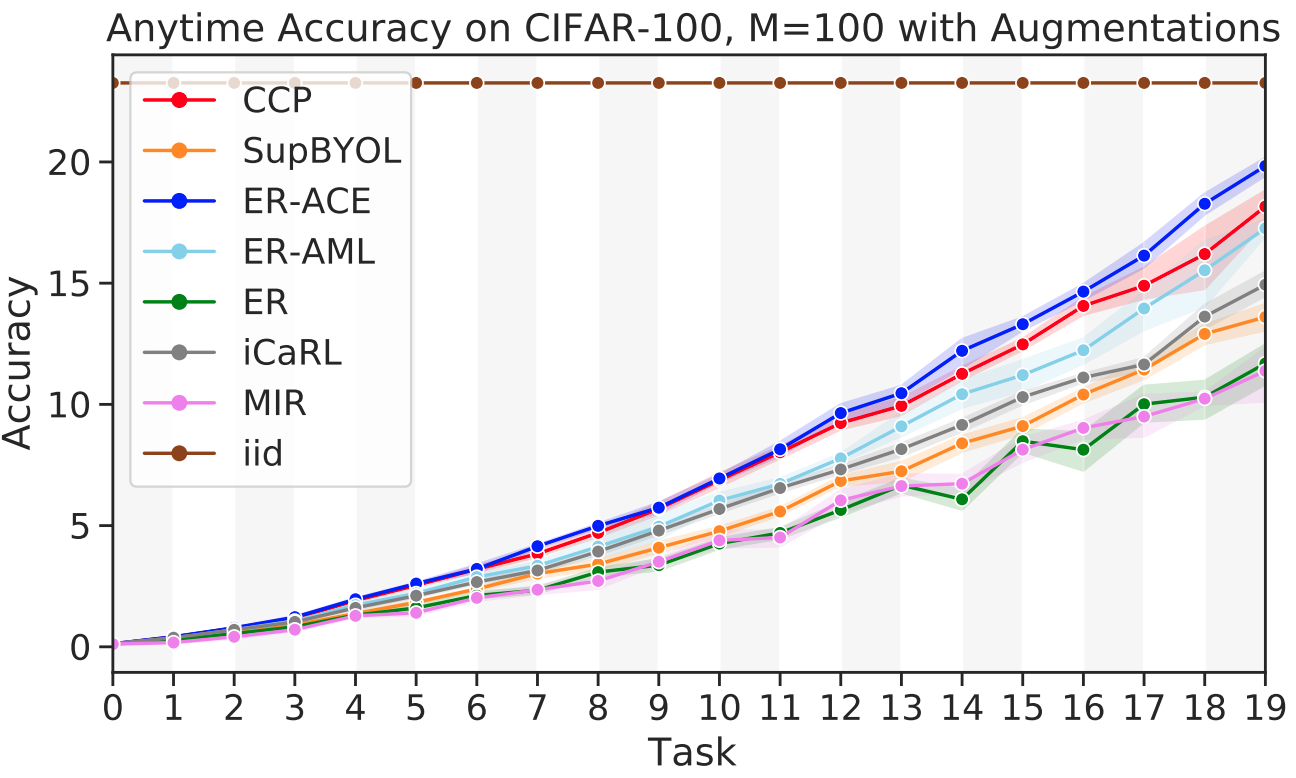}
    \end{minipage}\hfill
\end{figure*}